\documentclass[sigconf]{acmart}
\AtBeginDocument{%
  \providecommand\BibTeX{{%
    \normalfont B\kern-0.5em{\scshape i\kern-0.25em b}\kern-0.8em\TeX}}}

\setcopyright{acmcopyright}
\copyrightyear{2018}
\acmYear{2018}
\acmDOI{10.1145/1122445.1122456}

\acmConference[Woodstock '18]{Woodstock '18: ACM Symposium on Neural
  Gaze Detection}{June 03--05, 2018}{Woodstock, NY}
\acmBooktitle{Woodstock '18: ACM Symposium on Neural Gaze Detection,
  June 03--05, 2018, Woodstock, NY}
\acmPrice{15.00}
\acmISBN{978-1-4503-XXXX-X/18/06}

\usepackage{graphicx}
\usepackage{amsmath}
\usepackage{array}
\usepackage{float}
\usepackage{algorithm}
\usepackage{mathtools} 
\usepackage{subcaption}
\usepackage{algpseudocode}
\usepackage{amsthm}
\algnewcommand{\LineComment}[1]{\State \(\triangleright\) #1}
\newtheorem{thm}{Theorem}
\begin{document}

\title{Bandit based centralized matching in two-sided markets for peer to peer lending}

% %%
% %% The "author" command and its associated commands are used to define
% %% the authors and their affiliations.
% %% Of note is the shared affiliation of the first two authors, and the
% %% "authornote" and "authornotemark" commands
% %% used to denote shared contribution to the research.
\author{Soumajyoti Sarkar}
\authornote{Work done while the author was at Arizona State University}
\email{sarkar.soumajyoti@gmail.com}

\begin{abstract}
Peer lending platforms allow sequential fundraising for projects where multiple investors can selectively choose to fund different projects at a given point in time. However, understanding what makes a good investment choice for a lender in the presence of investment constraints has been an open ended question. The centralized investment mechanism in these platforms makes it difficult to understand the implicit competition that borrowers face from a single lender at any point in time. Matching markets are a model of pairing agents where  the preferences of agents from both sides in terms of their preferred pairing for transactions can allow to decentralize the market. We study investment designs in two sided platforms using matching markets when the investors or lenders are also subject to constraints on the investments based on borrower preferences. This situation creates an implicit competition among the lenders in addition to the existing borrower competition, especially when the lenders are uncertain about their standing in the market and therefore whether their preferences for projects are a good investment choice. We devise a technique based on sequential decision making that allows the lenders to adjust their choices and thereby allows them to learn feasible choices that could be good investments for them. We simulate two sided market matching in such a sequential decision framework using multi-armed bandits and a matching algorithm. We show the dynamics of the lender regret amassed compared to one scenario of an optimal borrower-lender matching. We observe  through simulations that the lender regret depends on the constraints of the matching objective and how the algorithm that executes the sequential decision aspect allows the lenders to learn and affects the overall regret dynamics of the system over a given time horizon.
\end{abstract}

% %%
% %% The code below is generated by the tool at http://dl.acm.org/ccs.cfm.
% %% Please copy and paste the code instead of the example below.



\maketitle

\section{Introduction}

%%quadratic funding 

Sequential decision making in two sided markets like consumers and producers has been part of bidding in e-commerce platforms like eBay, eBid for a very long time. Not only that, P2P platforms like Prosper in the past allowed lenders to bid on projects for peer microlending until they switched to posted price mechanism \cite{ceyhan2011dynamics}. However, for most peer microlending platforms like Kiva, LendingClub among others, sequential decision making among lenders is obscure. On the other hand single investor organizations like venture capitals have a monopoly on who they fund \footnote{https://www.inc.com/christine-lagorio/sam-altman-yc-monopoly.html} and this eludes any competition among multiple lenders over borrowers. In this paper, we therefore attempt at abstracting away peer lending models using matching markets. There are three keys points we address in this paper: first, matching markets assume that the agents know their preferences over each other \cite{rastegari2016preference}, however in peer lending platforms, the lender investments depend on a lot of factors like the return on investment, lender portfolio interests, uncertainty on borrower project success \cite{zhao2016portfolio}. To overcome this assumption, we introduce sequential decision making so as to allow lenders to adjust their preferences by learning them over time. 

Secondly, the implicit nature of matching causes competition among agents on both side - the borrowers trying to raise money from the same set of investors while the investors trying to invest in the selected projects. This nature of competition in markets for resolving conflicts has been studied recently \cite{liu2020competing} where the agent preferences on one side are concealed from the other and so the sequential decision aspect comes into play for preference revisions over time.  As mentioned, in such P2P platforms, there are mainly two sides to the market: the borrowers who want to borrow money from others for their projects or startups and the lenders  who lend money to borrowers. The traditional rule involves in such two sided trading follow the Dutch Auction Mechanism \cite{kumar1998internet,wei2017market}. However, we assume that the matching in our case is not dependent solely on the highest amount an investor wants to put in on a project, and there is no bidding from lenders in that each lender has a fixed budget which is the same for all borrower projects Therefore, we also assume there is no incentive for strategic truthfulness or concealed payments.

Finally, in an online setting, the learning step involves allowing the lender to adjust its utility or its valuation for a given project over multiple rounds. The optimization for matching the users depends on the submitted lender utilities. However, this optimization of utilities can be different from the rewards that the system provides to the lender on a successful match although they can be indirectly tied in the optimization. There have been recent studies that try to address this gap between what the lender estimates and what it receives as reward in an online setting \cite{johari2021matching}. We devise a mechanism to tie the rewards from the agents to their utility in an attempt to understand the dynamics of regret over time.   Centralized platforms to tackle these issues could ensure that the transactions between borrowers and lenders are not only based on the money that an investor is willing to put and its preferences but a borrower's willingness to accept the investment (these could be due to issues in lender terms\footnote{https://siliconhillslawyer.com/2019/03/03/standard-term-sheets-problem-yc/} or borrower's assessment of the investor profile). As an added caveat, it also allows for potential bias mitigation that can be implicit in such platforms \cite{sarkar2020mitigating}.

From the lender's perspective, the main signals of interest for utilities from borrower projects then constitute the probability of winning the bid, the probability of the loan being fully funded, as well as the returns from the investment. In keeping with these expectations, often the borrower's interests are sidelined as it is assumed that its only expectation from the platform and from investors is to get its project or startup funded. For the rest of the paper, we lay the foundations of our work that demonstrates a way to address competition and fair play in such peer lending platforms with ideas from matching markets\cite{roth1993stable}. The rest of the paper discusses some choices that could be made towards formulating the utilities on both sides, the mechanism for sequential decision making over rounds, and finally the tradeoff between the preference revisions and the rewards for the agents which are also tied in some ways. Throughout the paper, we consider agents are not strategic and therefore their preference submissions are honest.

\section{Related Work}
In this section, we start by laying out the motivations behind the research conducted in this study and the several studies done previously that are closely tied to our problem. The problem we study has been split evenly in the economics literature as well as the computer science discipline and we describe the related studies in three categories: \\

\noindent \textbf{1. Peer lending in markets:} Peer lending has been studied for platforms like Prosper which in its earlier days would allow users to bid on the projects. Earlier studies on this was conducted in \cite{chen2014auctions, chen2009social, chen2011market} where algorithms were designed for allocations of the social lending market that ensure stability and Pareto optimality consider equitable allocations among equal borrowers. In these studies, characterization of the Nash equilibria of the allocation mechanism was also conducted and disparities among the allocations with respect to borrower repayments. In our paper, we mainly focus on the uncertainty aspect of the market lenders and we try to understand the lender regrets when we optimize for lender and borrower utilities which can change over time unlike the studies mentioned. Dynamic Matching Market Design catering to such a context has been an area that has received attention in the past with studies \cite{akbarpour2014dynamic} characterizing dynamic matching in networked markets, where agents arrive and depart stochastically. One of the main motivations behind this work has been the recent work conducted in \cite{liu2020competing} which designed the notion of competition between agents in a matching market especially when one side of the market is uncertain about its preferences on the other side of the market. There have been other studies extending this framework especially in \cite{sankararaman2020dominate} which consider uniform valuation in demand side agents in the market and propose a decentralized version of the market that does not require knowledge of the time horizon or the suboptimality gaps for the system to be in equilibrium or the agents to reach stable matching. In our study, we consider different settings than considered in these studies in that firstly we do not consider global rankings of demand side agents among the supplier side agents. Instead all agents on one side have unique preferences for the agents on the other side. In this view, it becomes more difficult to design algorithms that can provably reach equilibrium. We consider the case of many-one matching where each lender is able to pick only one arm at each round. However, this model can be extended to the situation of many-many matching where each lender is able to pull multiple arms at each round.  Our model is closely related to the study in \cite{nguyen2021stability} of many-to-one matching markets in which agents with multi-unit demand aim to maximize a cardinal linear objective subject to multidimensional knapsack constraints.\\

\noindent \textbf{2. Bandits and bidding:} The idea of decision making under uncertainty for matching agents with budgets and preferences enjoy a rich literature in the field of multi-armed bandit (MAB) settings. Our model for adapting this framework of competing agents in matching markets for lending is motivated in part by the idea of dynamic pricing with limited supply \cite{babaioff2015dynamic} where we  we may have multiple products for sale, with a limited supply of each product. There have been studies that have proposed solving this problem in the realm of knapsack settings \cite{agrawal2014bandits, badanidiyuru2013bandits}. In these settings, at each round, the agents consume some resources as outcomes and get a reward which are accumulated over time. However in our settings, we assume the general MAB scenario where each round is a new matching and the agents learn their preferences over the rounds instead of allocations which are disbursed over time. Our settings are more closely aligned to ad allocation with budgets which have several studies associated to the exploration and exploitation settings of the ad allocation mechanism that considers the rewards allocated to advertisers\cite{gonen2007incentive, combes2015bandits}. The second area where our work is related in the field of bandits and auctions is the area of multi player bandits where the utilities of players impact the decision outcomes.  Recently, there have been studies in the field of multi-layer bandit settings \cite{bistritz2020my} in which the reward of a player is a stochastic function of the decisions of other players that operate in the same environment. The goal there is to design a distributed algorithm that learns the matching between players and arms that achieves max-min fairness while minimizing the regret. Such applications of fairness constraints in the choice of arms have also been considered in \cite{joseph2016fairness} where fairness precludes a worse agent being never favored over a better one, despite a learning algorithm's uncertainty over the true payoffs. Fairness in two sided markets includes settings where producer-consumer allocations are jointly optimized \cite{patro2020fairrec,chen2020reducing}. We consider a version of fairness in our work where we minimize some objective capturing the discrepancies in  allocation of resources among different borrowers.\\

\noindent \textbf{3. Recommendation systems and matchings}: Bandits have been used in recommendation systems for crowdfunding before. In a recent work on firing bandits \cite{jain2018firing}, the authors demonstrate a way to recommend projects to investors in a way that maximizes the number of projects that reach the funding goal. However, these studies do not take into account the preferences of the agents while proposing the recommendation strategy. One of the problems that come along with recommendation systems and their abstraction with matching markets \cite{tu2014online,chen2019prediction} is the gap between the utility of the agents and the regret that comes with the optimization objective of the recommendation systems. To this end, recent studies \cite{chiesa2014bridging} have focused on formal studies understanding the gap between utility maximization and regret minimization. In our work, we tackle this challenge by connecting the utilities of the agents with the regret the system computes over time. Utility based constrained matching optimization for recommendation systems have been a subject of research \cite{mladenov2020optimizing,zou2019reinforcement} where the formulation includes the constraints of consumers needing to maintain certain levels of engagement to stay in the system. Finally, user preferences in recommendation systems have been a key factor behind their successes and exploitations in item recommendations create a bias feedback loop \cite{schmit2018human} which can be solved using sequential decision making settings.

\begin{table}[!t]
	\centering
	\renewcommand{\arraystretch}{1}
	\caption{Table of Symbols}
	\begin{tabular}{|p{1.7cm}|p{6cm}|}
		\hline 
		{\bf Symbol} & {\bf Description}\\ 
		\hline\hline
		$b$($\mathcal{B}$), $l$($\mathcal{L}$)          & borrower (set of of borrowers), lender (set of lenders)\\
		\hline
		$c_b$, $q_l$ & borrower $b$'s request amount, lender $l$'s budget
			\\
		\hline
		$u_b(l), u_l(b)$ & borrower $b$'s utility from lender $l$, $l$'s utility from $b$
		\\
		\hline
		$\mathbf{Z}$ & a matching such that $z_{bl}$=1 if $b$ is matched to $l$
		\\
		\hline
		$m_z(l)$, $m_z(b)$ & denotes the borrower $\in \mathcal{B}$ matched to $l$, set of lenders matched to $b$
		\\
		\hline
		$t$ & time step
		\\
		\hline
		$\mathcal{M}_{\mathcal{L}}$, $\mathcal{M}_{\mathcal{B}}$ & dictionary mapping lenders to borrowers, dictionary mapping borrowers to sets of lenders
		\\
		\hline
		$X_{l, t}(b)$ & random outcome or reward of the lender $l$ from $b$ at time step $t$
		\\
		\hline
		$\mu_l(b)$ & empirical mean of the $b$-$l$ pair utility including the $u_l$ and borrower rewards $X_l$
		\\
		\hline
		$T_{b, l}(t)$ & the number of times borrower $b$ was matched to lender $l$ till time $t$.
		\\
		\hline
		$cu_l(b), cl_l(b)$ & Upper Confidence Bound (UCB) estimate of the lender utility $l$ from borrower $b$, Lower Confidence Bound for the same.
		\\
		\hline
	\end{tabular}
	\label{tab:table0}
\end{table}
\section{Problem Statement and Technical preliminaries}

Our model of lending through a market matching perspective is very close to the Shapley-Shubik model of bilateral trade with indivisible goods \cite{shapley1971assignment} where there is a set of buyers or bidders (the lenders in our case) and a set of sellers selling a unit of good (borrowers in our case) and no lender wants more than one unit of the good.  There is a monetary value that a buyer assigns to the seller's good and this relates to the amount of money that a lender is willing to lend to a borrower posting in our case despite what the borrower project funding requirements (which are generally more than an individual lender can contribute) are. In the rest of the paper, we will denote matrices using bold capital symbols, and vectors with bold lower cased symbols. A summary of the important symbols can be references in Table~\ref{tab:table0}. \\

\noindent \textbf{Borrowers and Lenders}: We model the lending platform as a market with 2 sides - the lenders denoted by the set of agents $\mathcal{L}$ = $\{l_1, l_2, \ldots l_N\}$ and the borrowers denoted by the set of arms $\mathcal{B}$ = $\{b_1, b_2, \ldots b_K\}$ and we assume that $K \leq N$. We now have a two-sided market where the agents or arms on the borrower side each have their  own funding request proposals and their corresponding requested amount which we denote by $c_b$, where $b \in \mathcal{B}$. Similarly, the lenders each have an overall budget $q_l$, where $l \in \mathcal{L}$. In addition, each set of agents on one side of the market have the opportunity to submit their preferred rankings of the agents on the other side of the market to the platform or the designer. These preferences can be conflicting - many lenders might prefer to lend to the same borrower, while multiple borrowers may prefer to tie up with the same lenders having specific portfolio and interests. \\

\noindent \textbf{Desiderata for Matching}: As mentioned above, we consider the case of many-one matching to simplify our settings similar to work done in \cite{bodine2011peer}, however our framework can be easily extended to the many-many setting albeit with more constraints. Each lender $l$ can be matched to at most one borrower while each borrower $b$ can be matched to multiple lenders based on the amount $c_b$ requested. Such mechanisms are currently followed in platforms like GoFundMe or Prosper Full Coverage lending model where a successful match denotes that borrower only gets the project funded when the sum of amounts lent, match or exceed the requested amount or a reserve price.

In our work, matching happens over multiple time steps and the lender is allowed to explore its options to realize its best matching over time. At each time step $t$, each lender $l_i$ is matched to a borrower $b_j$ and the lender receives a stochastic valued reward $\hat{\mu}_{b}(l)$ $\in (0, 1)$ independent of all other rewards that other lenders receive. The borrower or  arm means $\mu_{l}(b)_{b \in \mathcal{B}, l \in \mathcal{L}}$ are heterogeneous across lenders. Additionally, we also assume that for each lender $l$, the set of mean $\mu_l(b)_{b \in \mathcal{B}}$ are all distinct. We maintain a utility matrix $\mathbf{U_b} = \{\mathbf{u_{b_1}}, \mathbf{u_{b_1}}, \ldots\}$ of dimension $K\times N$ that stores each arm's utility from the lenders. Similarly we maintain another matrix $\mathbf{U_l}$ of dimension $N\times K$ that stores the lender's utilities from the borrowers. One of the significant points to note here is the gap between the lender's own utility estimate $u_l(b)$ for a borrower and the reward it receives $\hat{\mu}_l(b)$ if it is matched at a time step. The goal of the borrower-lender matching is to find an assignment of $m_{l, t} \in \mathcal{B}$ for all lenders $l \in \mathcal{L}$ at each $t$ pertaining to constraints based on these utilities and rewards which will be described in details in the next sections. We will often drop the time step symbol $t$ from the notations when we generalize the operations for all time steps.

\section{Key Ideas and intuition}

In terms of the market design for two sided lending, we first describe the matching objective and the utilities associated with the borrowers and lenders that go into their decision making.

\subsection{Matching objective}
In lieu with the above desiderata, the matching platform takes as input the ranking preferences of the borrowers and the lenders in the form of the utility functions of the borrowers and the lenders prior to each time step. At each time step, the platform solves a multi-objective optimization problem that aims at matching the borrowers and the lenders. To decide a matching between $\mathcal{B}$ and $\mathcal{L}$, we introduce the binary decision variable $\mathbf{Z}$ := $(z_{bl})_{(b, l)\in \mathcal{B} \times \mathcal{L}}$ such that $z_{bl}$ = 1 if the loan from lender $l$ is assigned and accepted by borrower $b$ and 0 otherwise. So  $\mathbf{Z} \in \{0, 1\}^{|\mathcal{B}| \times |\mathcal{L}|}$ is a matching. This gives rise to a policy that automatically matches a borrower and lender based on this optimization problem and the lender $l$ receives the reward $X_{l, t}$ for being matched to borrower $b$ at $t$. We will discuss about the reward structure $X_{l, t}$ in the following sections.  In our bandit setting, the lender $l$ is unable to observe the outcomes of other lenders nor the lenders matched to $b$.  To simplify  our settings, we make the following assumptions:

\begin{enumerate}
    \item All utilities $u_b(l)$ and $u_l(b)$ are non-negative $\forall b \in \mathcal{B}, l \in \mathcal{L}$.
    \item for a lender $l$, $u_b(l)$ $\neq$ $u_{b'}(l)$, for $b \neq b'$ and similarly, for a borrower $b$, $u_l(b)$ $\neq$ $u_{l'}(b)$, for $l \neq l'$.
\end{enumerate}

The borrower-lender pair $(b, l)$ yields a total utility of $u_{bl}:= u_b(l) + u_l(b) $. Recall that we consider a many-one matching where each borrower is matched to many lenders and each lender is matched to only one borrower. The preference orders of the borrowers and the lenders can be captured in the following way: $b \succ_j b' \Longleftrightarrow u_j(b) >  u_j(b')$ and $l \succ_i l' \Longleftrightarrow u_i(l) >  u_i(l')$. For a matching $\mathbf{Z}$, let $m_{l, t}\in \mathcal{B}$ be the borrower assigned to lender $l$ at time $t$, and $\mathbf{m}_{b, t} \subseteq \mathcal{L}$ be the set of lenders that are assigned to borrower $b$ at time $t$, that is, $m_l = b \Longleftrightarrow z_{bl} = 1$, and $\mathbf{m}_b := \{l \in \mathcal{L} \ | \ z_{bl} = 1\}$, considering the notations by dropping $t$ when we generalize the above for any step $t$. In this parlance, a pair $(b, l) \in \mathcal{B} \times \mathcal{L}$ is a blocking pair for $\mathbf{Z}$ if the following conditions \textbf{C1} are satisfied:

\begin{enumerate}
    \item $z_{bl}=0$
    \item $z_{b'l}= 0$, $\forall b'\neq b, b' \in \mathcal{B}$ or $u_l(b)$ $>$ $u_{l}(m_l)$
    \item $\sum_{l \in \mathbf{m}_b} c_l$ $< q_b$ or $\exists$ $l' \in \mathbf{m}_b$ such that $u_b(l) > u_b(l')$.
\end{enumerate} 
\vspace*{.1in}

A stable matching is defined based on the absence of blocking pairs. The stable matching model introduced by Gale-Shapley \cite{shapley1971assignment} finds a matching using the deferred acceptance (DA) algorithm. However it does not implicitly consider the notion of utility between the agents in that the utiltity values are concealed in the preference rankings of the agents. Another commonly employed algorithm is Gales’ top trading cycles (TTC) algorithm \cite{shapley1974cores}, produces a matching that is Pareto-efficient but not necessarily stable. On the other hand, as described in the previous section, the Gale-Shubik model maximizes the total utility gained by the agents on either side of the market based on the assumption that the utility can be exchanged between a borrower-lender pair in the matching. This assignment model can be reduced to a linear optimization problem but these do not maximize the total utility or agent specific utility and suggests that the notion of stability and the objective of utility are in general incompatible. To this end, we formulate our objective of both stability respecting the preferences of the borrowers and lenders while maximizing lender returns using a linear program formulation. The Gale-Shapley notion of stable matching considering lender budget $q_l$ and borrower request $c_b$ for a pair $b-l$ can be characterized by the following linear inequality as mentioned in \cite{baiou2000stable}:

\begin{equation} 
    c_b z_{bl} + c_b\sum_{b' \succ_l b }z_{b'l} + \sum_{l' \succ_{b} l} q_{l'}z_{bl'}  \geq c_b
\end{equation} \label{eq:gs}
The proof of the statement can be explained in the following way.
\begin{thm}
Suppose that  $\mathbf{Z} \in \{0, 1\}^{|\mathcal{B}| \times |\mathcal{L}|}$ is a matching. Then, it is stable if and only if constraint \ref{eq:gs} is satisfied.
\end{thm}

\noindent \textbf{Proof:}
We prove the theorem by contraposition. If constraint \ref{eq:gs} is violated then it implies $\exists (b, l') \in \mathcal{B} \times \mathcal{L}; \ z_{bl} = 0, \ \sum_{b' \succ_l b }z_{b'l}=0$, and $\sum_{l' \succ_{b} l} q_{l'}z_{bl'}  < c_b$ which $	\Longleftrightarrow$ $\exists (b, l') \in \mathcal{B} \times \mathcal{L}$; $(b, l')$ is a blocking pair for $\mathbf{z}$ following the assumptions \textbf{C1}. This in turn implies \textbf{Z} is not stable.

We use this notation to arrive at a linear program that maximizes the utility of the matching while minimizing the number of blocking pairs. Defining a binary decision variable $\mathbf{w}$ := $(w_{b, l})$ $\in \{0, 1\}^{|\mathcal{B}| \times |\mathcal{L}|}$, we use the following constraint: 

\begin{equation}
        c_b z_{bl} + c_b\sum_{b' \succ_l b }z_{b'l} + \sum_{l' \succ_{b} l} q_{l'}z_{bl'}  \geq c_b (1- w_{bl})
\end{equation} 

This characterization of the stability of matching markets can be understood using the following observation: when $w$=0, this inequality is the same as constraint \ref{eq:gs}, and so $w_{bl}$ denotes that $(b, l)$ is a blocking pair and so the number of blocking pairs is given by  $\sum_{b} \sum_{l} w_{bl}$. The objective is to maximize the utility and minimize the number of blocking pairs at the same time. The matching objective keeping the above constraints can be formulated as \textbf{MQ1}:

\begin{equation} \label{eq:gs_opt}
\begin{array}{ll@{}ll}
\text{maximize } \quad &  \lambda_1\sum_{b\in \mathcal{B}} \sum_{l \in \mathcal{L}}u_l(b) x_{bl}   - \lambda_2 \sum_{b\in \mathcal{B}} \sum_{l \in \mathcal{L}} w_{bl} 
    \end{array}
\end{equation}
\[
\begin{array}{ll@{}ll}
\text{subject \  to} & \sum_{b\in \mathcal{B}} x_{bl} \leq 1  \ \ \ \ (\forall l \in \mathcal{L})\\
& \sum_{l\in \mathcal{L}} x_{bl}q_l \geq c_b \ \ \ \  \ (\forall b \in \mathcal{B}) \\
& c_b x_{bl} + c_b\sum_{b' \succ_l b }x_{b'l} + \sum_{l' \succ_{b} l} q_{l'}x_{bl'} \\ & \geq c_b (1- w_{bl})  (\forall l \in \mathcal{L}, \forall b \in \mathcal{B})\\
& x_{bl} \in \{0, 1\} \ \ \ \ ( \forall b \in \mathcal{B}, \forall l \in \mathcal{L}) \\
& w_{bl} \in \{0, 1\} \ \ \ \ ( \forall b \in \mathcal{B}, \forall l \in \mathcal{L}) 
\end{array}
\]

We denote the set of constraints above as \textbf{C2}. Briefly these constraints satisfy the following: (1) the lenders can only be matched to one borrower, (2) the number of blocking pairs (denoted by $w_{bl}$) should be minimized in accordance with the original stable matching constraints \cite{roth1993stable}, and the borrower's requested amount must exceed the sum of investments from matched lenders. This is in addition to the constraint \ref{eq:gs} which minimizes the number of blocking pairs while maximizing the utility. Note above that we optimize for the lender utility in Equation~\ref{eq:gs_opt} but we will come back to this setting when we evaluate our matching objective and which also constitutes the need for our IP formulation instead of the traditional Gale-Shapley agent optimal algorithm.

\subsection{Utility constraints}
Reasons for lender preferences over borrowers could arise from the return on investment (ROI) which could be calculated in a myriad ways using a lot of other factors \footnote{http://blog.lendingrobot.com/research/calculating-financial-returns-in-peer-lending/}. For the borrower, the main reason to prefer one lender over another is the past reputation of the lender (since network effects can significantly accelerate the funding \cite{horvat2015network}) as well as the interest matches (especially in VC funding, the investor liquidation preferences can play a role in startup preferences). As for the lender case, we sample $u_b(l)$ from a uniform distribution.  These utilities have been optimized for lender returns in settings of recommendation systems \cite{choo2014understanding}. In our settings for preference elicitation, the matching objective function with constraints as defined in the previous section depends on the preferences that are set by agents on both sides at each step of the matching. We consider the utilities $\mathbf{u}_b = \{u_b(l_1), u_b(l_2)...\}$, $l_i \in \mathcal{L}$ and $\mathbf{u}_l= \{u_l(b_1), u_l(b_2)...\}$, $b_i \in \mathcal{B}$ that denote the vector of values for each agent $l$ or $b$ about its preferences of agents on the other side of the market. And as stated before, the ordering depends on the value that the agents estimate prior to matching. There are two ways we use these utilities in our market design:

\begin{itemize}
    \item \textbf{Lender utilities} - Preferences from each lender are elicited using their utility functions. So, each lender $l$ ranks the borrowers using their utilities set at the start of the matching. However, in the sequential decision time step, each lender gets to revise its utilities based on the rewards $X_{l, t}(b)$ it obtains from the matching at a particular round. In our setting, the lenders adjust their preferences for the borrowers based on $X_{l, t}(b)$ if $m_l=b$ at time step $t$ for matching $z$. We will discuss the choices of $X$ in Section~\ref{sec:seq_dec} to ensure calibration of the utility values. In our work we consider that $u_l(b)$ depends on two factors: lender budget $q_l$ and borrower rate $\eta_b$/.
    
    \item \textbf{Borrower utilities} - Similar to the lender, each borrower $b$ ranks the borrowers using their utilities set at the start of the matching. However, unlike the lender, we assume that the borrower utilities are fixed over time.
\end{itemize}

\section{Sequential decision making} \label{sec:seq_dec}
One important point to recall is that the agents on each side are not aware of the preferences of each other irrespective of which side they belong to, which is why the case for competition arises more prominently. In hindsight, such preferences are not globally known since user preferences for each other change over  time,  so the estimated utilities also change over time . In order to arrive at a preferred matching faster, we operationalize the matching platform with sequential decision making in the form of multi-armed bandits with a centralized matching platform. This notion of centralized matching markets has been studied before in \cite{liu2020competing, bistritz2020my}. In the framework of  sequential decision making, we allow lenders to submit their preferences over multiple time steps and subsequently the matching happens in these steps or rounds. The lender receives a reward at each step determined by which borrower it gets matched to. From the lender's point of decision making, the uncertainty comes from the absence of knowledge of the borrower preferences (or utilities) to the lenders.  This is a bandit setting \cite{das2005two} where at each round, the platform provides a pseudo-reward  to the lender based on the borrower it is matched to and allows the lender to revise its preference rankings for the next round. The goal of each lender is to get matched to its most preferred lender at each round and this happens when the gap between the lender utility and the rewards narrows.

\begin{algorithm}[!t]
\caption{Matching between borrowers and lenders (GS-UCB)}
\label{alg:edge_infer}
\begin{algorithmic}[1]
\State 	\textbf{Input:} $\mathcal{B}$, $\mathcal{L}$,  $\mathbf{u}_l$, $\mathbf{u}_b$, $T$. 
\State \textbf{Output:}  Matching  dictionaries $\mathcal{M}_{\mathcal{B}}$, $\mathcal{M}_{\mathcal{L}}$ 
	\State $cu_l(b)$ $\leftarrow$ $\infty$ ($\forall b \in \mathcal{B}, \forall l \in \mathcal{L}$) 
    \State $T_{b, l}(0)$ $\leftarrow$ 0 ($\forall b \in \mathcal{B}, \forall l \in \mathcal{L}$) 

	\For {t = 1, 2, $\ldots$ T}   	
	    \State $\mathcal{M}_{\mathcal{B}}$, $\mathcal{M}_{\mathcal{L}}$  $\leftarrow$ Matching \textbf{MQ1} using $cu_l$ and $u_b$ ($\forall b \in \mathcal{B}, \forall l \in \mathcal{L}$)
		\For {l $\in$ $\mathcal{L}$}
			\State $m_{l, t}$ $\leftarrow$ $\mathcal{M}_\mathcal{L}(l)$  \Comment{matched borrower} 
    		 \If{$m_t(l)$ is not empty}
		        \State $r$ $\leftarrow$  $X_{l, t}(m_{l, t})$   \Comment{lender reward}
		        \State Update $\hat{\mu}_l(m_{l, t})$ using $r$ in Equation~\ref{eq:update_eq} 
                \State $T_{m_{l, t}, l} (t)$ $\leftarrow$ $T_{m_{l, t}, l} (t-1)$ + 1
		    \EndIf
            \For{b $\in \mathcal{B}$} 
                 \State $cu_l(b)$ $\leftarrow$ $\hat{\mu}_{l}(b)$ + $\sqrt{\frac{3 \ \mbox{log} \ t}{2 \ T_{b, l} (t)}}$ 
            \EndFor
        \EndFor
	\EndFor
	\State return  $\mathcal{M}_{\mathcal{B}}$, $\mathcal{M}_{\mathcal{L}}$ 
\end{algorithmic}
\end{algorithm}

\subsection{Matching with UCB (GS-UCB)}
In this setting we assume that the horizon T is not known to the agents and is much larger than $|\mathcal{B}|$ and $|\mathcal{L}|$, since we assume the matching happens over a long period of time.
In what follows, we explain how the reward distributions for each lender are calculated and which lays the path for the exploration of the arms (here the borrowers) by the lenders at each round. Before describing the desiderata for the reward structure, we recall that for a lender $l$, the set of arms or borrowers $b \in \mathcal{B}$ are associated with random variables $X_{l, t}(b)$, $t \geq 1$ with bounded support on [0, 1]. The variable $X_{l, t}(b)$ indicates the random outcome of the borrower $b$ at time $t$ such that $m_t(l)=b$. These set of  random variables $\{X_{l, t}(b) \  | \  t \geq 1\}$ associated with lender $l$ are independent and identically distributed according to some unknown distribution with expectation $u_b(l)$, $\forall b \in \mathcal{B}$.  The empirical mean for the lender $l$ for the set of borrowers is denoted by $\hat{\mu}_{l}(b)$.

Now, we describe the matching algorithm utilizing a popularly technique in the set of bandit based algorithms known as the Upper Confidence Bound (UCB) \cite{lai1985asymptotically}. At each time step in Algorithm~\ref{alg:edge_infer},  the platform matches lender $l$ with borrower $m_l$ upon which $l$ is deemed to be able to pull the arm successfully and gets to know the reward $X_{l, t}(b)$. Lender $l$ updates their empirical mean $\hat{\mu}_{l}(m_l)$ through the following equation: 
\begin{equation}\label{eq:update_eq}
  \hat{\mu}_{l}(b) = u_l(b) + \frac{1}{1+T_{b, l}(t)} \Big[\sum_{s=1}^t \mathbf{1} \{m_s(l) == b\} X_{l, s}(b) \Big] 
\end{equation} 

where $T_{b, l}(t)$ = $\sum_{s=1}^t \mathbf{1}\{m_l(s) == b\}$ is the number of times borrower $b$ was matched to lender $l$ till time $t$. In Algorithm~\ref{alg:edge_infer}, $\mathcal{M}_l$  denotes the storage structure mapping each borrower to a lender $l$ and similarly and $\mathcal{M}_b$ denotes the set of lenders matched to a borrower which is output at each time step by \textbf{MQ1}.  We utilize the Upper Confidence Bound (UCB) design \cite{lai1985asymptotically} where at each time step $t$ the lenders compute the upper confidence bound for each borrower as follows:

\begin{equation} \label{eq:cub}
cu_{l}(b) = 
\begin{cases}
\infty \ \ \ \ \ \ \ \ \  \ \ \ \ \  \ \ \ \ \ \ \ \ \ \ \ \ \ \  , \ T_{b, l}(t) = 0 \\
 \hat{\mu}_{l}(b) + \sqrt{\frac{3 \ \mbox{log} \ t}{ 2 \ T_{b, l} (t)}} \ \ \ \ \ \ \ \ , \ otherwise \\
\end{cases}
\end{equation}

Each lender $l$ ranks the arms $b$ according to $cu_l(b)$ and sends the new utilities $\mu_l(b)$ to the platform while the borrower preferences remain unchanged. The important point to note is that at each step from $t \geq 1$, the objective utility to maximize for the lender is $\mu_l(b)$, however the preference ordering in the set of constraints \textbf{C2} is done using $\mathbf{cu}_l$.  \\

\begin{algorithm}[!t]
	\caption{Matching between borrowers and lenders (GS-BLEMET)}
	\label{alg:blemet}
\begin{algorithmic}[1]
    \State \textbf{Input:} $\mathcal{B}_{unm}$, $\mathcal{L}_{unm}$,  $\mathbf{c}$, $\mathbf{u}_l$, $\mathbf{u}_b$, $\mathbf{q}$.
    \State \textbf{Output:} Matching  dictionaries $\mathcal{M}_{\mathcal{B}}$, $\mathcal{M}_{\mathcal{L}}$
	\State $cu_l(b)$ $\leftarrow$ $\infty$ ($\forall b \in \mathcal{B}_{unm}, \forall l \in \mathcal{L}_{unm}$) 
	\State $cl_l(b)$ $\leftarrow$ $-\infty$ ($\forall b \in \mathcal{B}_{unm}, \forall l \in \mathcal{L}_{unm}$)  
    \State $T_{b, l}(0)$ $\leftarrow$ 0  ($\forall b \in \mathcal{B}_{unm}, \forall l \in \mathcal{L}_{unm}$) 
    \State $\mathcal{M'}_{\mathcal{B}}$ $\leftarrow$ $\{\}$, $\mathcal{M'}_{\mathcal{L}}$ $\leftarrow$ $\{\}$  ($\forall b \in \mathcal{B}_{unm}, \forall l \in \mathcal{L}_{unm}$) 

	\For{t = 1, 2, \ldots T}    		
       \State $\mathcal{M}_{\mathcal{B}}$, $\mathcal{M}_{\mathcal{L}}$  $\leftarrow$ Matching \textbf{MQ1} using $cu_l$ and $u_b$ ($\forall b \in \mathcal{B}_{unm}, \forall l \in \mathcal{L}_{unm}$)
		\For{b $\in$ $\mathcal{B}_{unm}$} 
	        \LineComment{Find all lenders matched to borrower $b$} 
		    \State $\mathbf{m}_b$ $\leftarrow$ $\mathcal{M}_{\mathcal{B}}$[$b$]  
		    \For{l $\in$ $\mathbf{m}_b$} 
		    	\LineComment{If  $b$ is currently $l$'s "most" preferred borrower} 
		        \If{$cl_{l}(b)$ $>$ $max \  \Theta_l(\mathcal{B}_{unm} \setminus b)$}
    		        \LineComment{If $l$ is most preferred among lenders for $b$} 
    		        \If {$cl_{l}(b)$ $>$ $max \ \Upsilon_b(p_n \setminus l)$ }
    		            \State $\mathcal{M'}_{\mathcal{B}}(b)=\mathcal{M'}_{\mathcal{B}}(b) \cup l$, $\mathcal{M'}_{\mathcal{L}}(l) = b$
    		            \State $\mathcal{L}_{unm}$ $\leftarrow$ $\mathcal{L}_{unm}$ $\setminus$ $l$ 
    		            \State $c_b$ $\leftarrow$ $c_b$ - $q_l$ 
    		        \EndIf
		        \Else
    		        \State $r$ $\leftarrow$  $X_{l, t}(b)$   \Comment{lender reward}
    		        \State Update $\hat{\mu}_{l, t}(b)$ using $r$ in Equation~\ref{eq:update_eq} 
		            \State $cu_{l, t}(b)$ $\leftarrow$ $\hat{\mu}_{l, t}(b)$ + $\sqrt{\frac{ 3 \ \mbox{log} \ t}{   2 \ T_{b, l} (t-1)}}$ 
		            \State $cl_{l, t}(b)$ $\leftarrow$ $\hat{\mu}_{l, t}(b)$ - $\sqrt{\frac{ 3 \ \mbox{log} \ t}{   2 \ T_{b, l} (t-1)}}$ 
		            \State Update $\Upsilon_b(l)$ and $\Theta_l(b)$ using $cu_{l, t}(b)$ and $cl_{l, t}(b)$ respectively 
	                \State $T_{b, l} (t)$ $\leftarrow$ $T_{b, l} (t-1)$ + 1
		        \EndIf
		    \EndFor
	    \EndFor
    	\For{b $\in$ $\mathcal{B}_{unm}$}
		    %\LineComment{If amount allocated to $b$ at time $t$ exceeds the requested amount} 
			\If{$C$[$b$] $\leq$ 0}
    			 \State $\mathcal{B}_{unm}$ $\leftarrow$ $\mathcal{B}_{unm}$ $\setminus$ $b$ 
			\EndIf
		\EndFor
	    \State Broadcast $\mathcal{B}_{unm}$ to the lenders $\mathcal{L}_{unm}$
	    \If{$\mathcal{L}_{unm}$ is empty or $\mathcal{B}_{unm}$ is empty}
	        \State break;
	   \EndIf
	\EndFor
	\State return $\mathcal{M'}_{\mathcal{B}}$, $\mathcal{M'}_{\mathcal{L}}$

\end{algorithmic}
\end{algorithm}

% \begin{thm}
% \noindent \textbf{ (Theorem 2 given by Chen et al. \cite{chen2013combinatorial})}  Consider a combinatorial MAB with an $(\alpha, \beta)$-approximation oracle. If the  bounded smoothness function $f(x) = \gamma . x^\omega$ for some $ \gamma > 0$ and $\omega \in (0, 1]$, the regret is at most:
% \end{thm}

% \begin{equation*}
%     \frac{2\gamma}{2-\omega} . (6m \ \mbox{ln} \ n)^{\frac{\omega}{2}}. n^{1-\frac{\omega}{2}} + \big( \frac{\pi^2}{3} + 1\big).m.\Delta_{max}
% \end{equation*}

\subsection{Matching with Early Termination (GS-BLEMET)}
One of the drawbacks of the baseline UCB algorithm is that in the absence of networked environments or absence of side observations as is the case with previous studies \cite{caron2012leveraging,mannor2011bandits}, the lenders are not able to view the preferences for lenders also competing for that borrower. This issue is also aggravated since we do not consider strategic agents in our settings and the preferences set by the lenders at each time step is bereft of external information about their standing among other lenders for a specific borrower. That is, if the lender $l$'s estiamte of the empirical upper bound $cu_l(b)$ of a borrower arm $b$ is already very high compared to the other lenders, and it gets matched to $b$ at $t$, then we can finalize the matching of $b$ to $l$, update $b$'s remaining amount to continue the matching for and remove the lender $l$ from the further rounds, so that the other lenders can revise their estimates of the borrowers. This is reminiscent of the successive elimination algorithm used in the bandit settings \cite{even2006action}. To that end, we devise an algorithm for \underline{B}orrower and \underline{LE}nder \underline{M}atching with \underline{E}arly \underline{T}ermination (BLEMET) described in Algorithm~\ref{alg:blemet}. We define two symbols: $\Upsilon_{b, t}$ =  $\{cu_{l}(b)\}$ $\forall l \in \mathcal{L}$, which stores the UCB of all lenders $l$ with respect to the borrower $b$ at step $t$, and $\Theta_l$ = $\{cu_{l}(b)\}$ $\forall b \in \mathcal{B}$ which stores the UCB of all borrowers $b$ with respect to the lender $l$'s estimates. The main steps in this design of the algorithm can be summarized in these three steps (note that in our centralized platform, the matching and decisions of termination are driven solely by the platform/matching algorithm itself and the agents drive this decision by adjusting their utilities through their preferences over time unlike a decentralized distributed setting):

\begin{enumerate}
    \item The goal is to match each lender to a borrower as soon as a criterion is met. To this end, we keep a list of unmatched borrowers and lenders at each time point as per our definitions of matching in $\mathcal{L}_{unm}$ and $\mathcal{B}_{unm}$ (these sets are initialized as $\mathcal{L}$ and $\mathcal{B}$ respectively). At each time step, we find the matching $\mathcal{M}_{\mathcal{B}_{unm}, \mathcal{L}_{unm}}$  using the optimization objective described in Equation~\ref{eq:gs_opt} to find the map of borrowers matched to lenders in Step 8. We follow the same initialization of $cu_b$ defined in Equation~\ref{eq:cub} and additionally, we define  $cl_b$ except we subtract the term in the square root in Equation~\ref{eq:cub} from $\hat{\mu}_l(b)$.
    
    \item For each borrower $b$ $\in \mathcal{B}_{unm}$, we extract the set of lenders $p_b$ who are matched to $b$ in the current time step. Lines 13-16 define the two criteria that we consider for early termination - if $l$'s estimate of LCB for $b$ is higher than the UCB of all other borrowers unmatched so far defined in Line 14 and if $l$'s LCB is higher than the UCB of all other lenders contesting for $b$ defined in Line 14. When both of these conditions are met, the borrower $b$ is set to be matched to the lender $l$ and is removed from further iterations and we update $\mathcal{L}_{unm}$ and borrower $b$'s remaining requested amount $c_b$. If either of the conditions are not met, then we just update the UCB and LCB estimates of the lenders $p_b$ for the borrower $b$, as shown in Lines 22-26. 
    
    \item The final step also includes checking whether a borrower has its request already met by the allocations so far after early termination. To this end, we check whether the borrower's remaining request  $c_b$ is less than 0, and in that case, we remove borrower $b$ from further iterations of the game. Finally, we broadcast the remaining borrowers to the lender. Line 37 is important in this algorithm in that the termination depends not just on the horizon T but on the UCb updates as well.
\end{enumerate}

We continue these three steps for all the iterations of the game until we reach the maximum iterations $T$ or all the borrowers and lenders have been matched, whichever happens earlier.

\section{Equitable Allocations}

So far, we have presented a situation that takes into account the best interests of both the borrowers and the lenders through their utility functions. However, as mentioned previously, there are several factors hat influence a lender's decision like the probability of a borrower listing getting fully funded, the probability of the lender getting matched, the interest rate \cite{zhao2016portfolio,ceyhan2011dynamics} which are uncertain, but apart from that there can be bias to the choices themselves - like preference for new listings, preference for specific countries \cite{burtch2014cultural}, preference for creditworthy borrowers, popularity bias as seen in recommendation systems \cite{abdollahpouri2019managing}. The added constraint of early termination in BLEMET can cause some borrowers to lose out on the bidding and to mitigate these issues, we add the following term that captures the worst case allocation to the borrowers. We want to minimize the maximum $c_{b, t}$ over time (note that $c_{b, t}$ is adjusted at each time step $t$ in GS-BLEMET) so as to ensure equitable allocations. We capture that by minimizing the following component in addition to the objectives optimized so fat:  $\kappa(t) = max_{b \in \mathcal{B}} \  c_{b, t} * e^{\omega \frac{t}{T}}$, where $\omega$ is a hyper parameter and  $t$ is the time step of the matching. Following this, we define the matching objective \textbf{M2} as:

\begin{equation}
\begin{array}{ll@{}ll}
\text{maximize } \quad &  \lambda_1\sum_{b\in \mathcal{B}} \sum_{l \in \mathcal{L}}u_{bl}z_{bl} \\   - \lambda_2 \sum_{b\in \mathcal{B}} \sum_{l \in \mathcal{L}} w_{bl} \\  - \lambda_3 \  \kappa(t)
    \end{array}
\end{equation}

\begin{equation*}
\begin{array}{ll@{}ll}
\text{subject \ \  to}  \ \ & \mathbf{C2}
\end{array}
\end{equation*}
We then execute the same steps  in Algorithm~\ref{alg:blemet} with the added constraints in Line 8. We refer to this algorithm as GS-BLEMET-FAIR.

\section{Experiments}

\subsection{Simulation Settings}
To understand how the  settings described in Section~\ref{sec:seq_dec} compare in terms of the regret dynamics, we simulate our findings with agents with utilities and rewards generated in a stochastic environment. We consider 20 borrowers and 60 lenders as the agents on two sides of the market. To assign a loan request budget for the borrowers and an investment budget for the lenders, we simulate using the following rule. We set the configurations of the simulations in the following way:

\begin{itemize}
    \item \textbf{Loans and Budgets}: We uniformly sample loans values $\mathbf{c}$ in the range of 10 and 50 for the borrowers. We fix the lender investment budgets $\mathbf{q}$ by selecting a value uniformly in the range of 1 and 30 such that the configuration of the borrower and lender budgets satisfy the following criterion: the sum of the lender amounts exceed the sum of the borrower requests. This configuration setting ensures that there would be a valid matching that could be obtained using the optimization objective satisfying the resource constraints of the lenders and the borrowers. However, this constraint could be relaxed even if we add the assumption that not all borrowers need to be matched in every time step. 
    
    \item \textbf{Utilities}: We randomly sample borrower rates $\boldsymbol\eta$ for each $b$ $\in \mathcal{B}$ from a uniform distribution in the range (0, 1) and the borrower utilities for each lender $u_b(l)$ are sampled from a uniform distribution in the range (0, 1). Following this, the lender utilities $u_l(b)$ are calculated as $\eta_b*q(l)$. We then normalize the lender utilities to calibrate them in the range of the borrower utilities.
    
    \item \textbf{Rewards}: As mentioned before, for each borrower-lender \textit{b-l} pair, we sample the rewards from a  1-subgaussian distribution with mean $u_b(l)$. One thing to note that these rewards $X_{l, t}(b)$ for a lender $l$ from $b$ at $t$ which depends on $u_b(l)$ would not necessarily be the same for each $t$ and this is where stochasticity comes into play. That is to say, the reward for a lender $l$ when matched to a borrower $b$ at time step $t$ could be different at time step $t'$ even if the lender $l$ was matched to $b$ both at $t$ and $t'$.
\end{itemize}

Before moving to the evaluation settings and the results, we discuss the hyperparameter values, especially $\lambda_1, \lambda_2$ and $\lambda_3$. For the setting without the fairness component in the objective, we set the $\lambda_1, \lambda_2$ to 0.5 each keeping them in equal proportions and for the objective including the fairness constraint, we set $\lambda_1$ to 0.5 and grid search for $\lambda_2$ and $\lambda_3$ such $\lambda_2 + \lambda_2$ = 0.5. One additional point to note is that since we consider early termination in the algorithm GS-BLEMET, when a lender $l$ is removed from the simulation instance at time step $t$, we consider the regret constant for all time steps after $t$ for lender $l$.

To simulate the matching, we run the different settings for 10000 steps and simulate each setting for 50 times. These 50 simulations are run keeping the initial utilities $\mathbf{u_b}$ and $\mathbf{u_l}$ unchanged as well as the hyper-parameters for the algorithm constant. The only stochastic components come from the uncertainty about the rewards obtained over time which control the learning mechanism.

% \begin{figure*}[!t]
% \centering
% \minipage{0.35\textwidth}
% \includegraphics[width=5.5cm, height=3cm]{lender_img21.png}
% \endminipage
% \minipage{0.3\textwidth}
% \includegraphics[width=5.5cm, height=3cm]{lender_img22.png}
% \endminipage
% \minipage{0.3\textwidth}
% \includegraphics[width=5.5cm, height=3cm]{lender_img3.png}
% \endminipage
% \hfill
% \\
% \minipage{0.35\textwidth}
% \includegraphics[width=5.5cm, height=3cm]{lender_img41.png}
% \endminipage
% \minipage{0.3\textwidth}
% \includegraphics[width=5.5cm, height=3cm]{lender_img48.png}
% \endminipage
% \minipage{0.3\textwidth}
% \includegraphics[width=5.5cm, height=3cm]{lender_img55.png}
% \endminipage
% \hfill
% \caption{Lender regret over time simulated over 50 runs.}
% \label{fig:loans_stats}
% \end{figure*}

\subsection{Evaluation}

To evaluate the quality of the lender assigned borrowers, we compute the following cumulative regret metric for each lender at each time step $t$:  $t*u_l(b_{opt}) - \sum_{i=1}^t \mathbb{E} \ \mathcal{R}(u_{l}(b_{alg}))$ where $\mathcal{R(.)}$ denotes the $X_{l}$ that the lender receives given that $b_{alg}$ is the matched borrower for lender $l$ at time $t$ returned in Algorithm~\ref{alg:edge_infer} while $b_{opt}$ is computed using the following optimization:  

\begin{equation} \label{eq:opt_fair}
\begin{array}{ll@{}ll}
\text{maximize } \quad & \lambda_1\sum_{b\in \mathcal{B}} \sum_{l \in \mathcal{L}}u_{b, l} z_{bl}   - \lambda_2 \sum_{b\in \mathcal{B}} \sum_{l \in \mathcal{L}} w_{bl} 
    \end{array}
\end{equation}
\[
\begin{array}{ll@{}ll}
\text{subject \  to} &  \textbf{C2} 
\end{array}
\]

Note we optimize the sum of the borrower and lender utilities as in hindsight, the lender would have adjusted its ranking based on borrower preferences had it have access to that information. However, we do not note that this setting we which call the \textit{optimal setting} is not guaranteed to produce the best matching in terms of the utility that the lender receives since this is the gap we have for some lenders attributed to the difference between the matching objective and the lender returns.

\subsection{Results}
We start by observing the results from the \textbf{GS-UCB} algorithm in Algorithm~\ref{alg:edge_infer}, and we observe three kinds of dynamics in the lender regrets:
\begin{figure}[!h]
\centering
\minipage{0.25\textwidth}
\includegraphics[width=4.5cm, height=3cm]{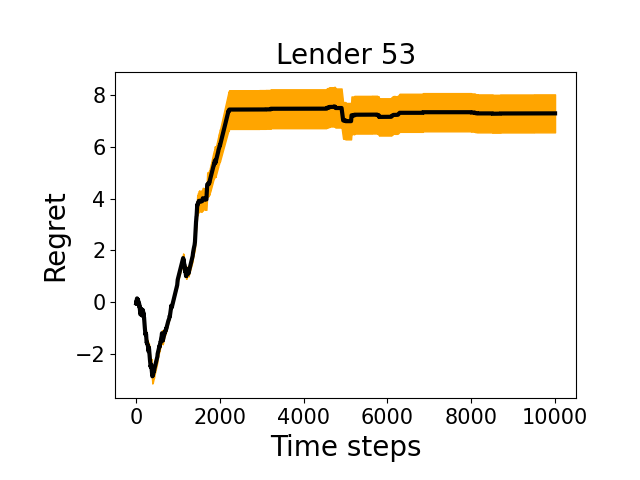}
\endminipage
\minipage{0.3\textwidth}
\includegraphics[width=4.5cm, height=3cm]{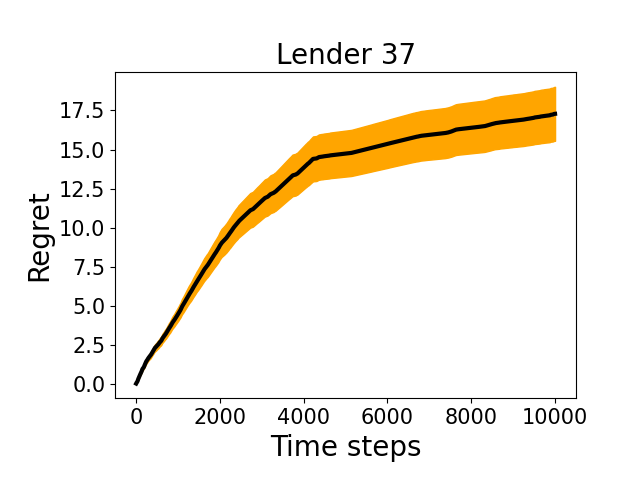}
\endminipage
\hfill
\\
\minipage{0.25\textwidth}
\includegraphics[width=4.5cm, height=3cm]{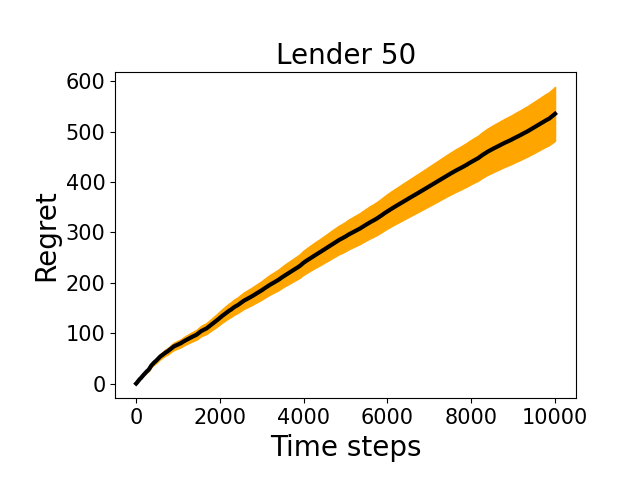}
\endminipage
\minipage{0.3\textwidth}
\includegraphics[width=4.5cm, height=3cm]{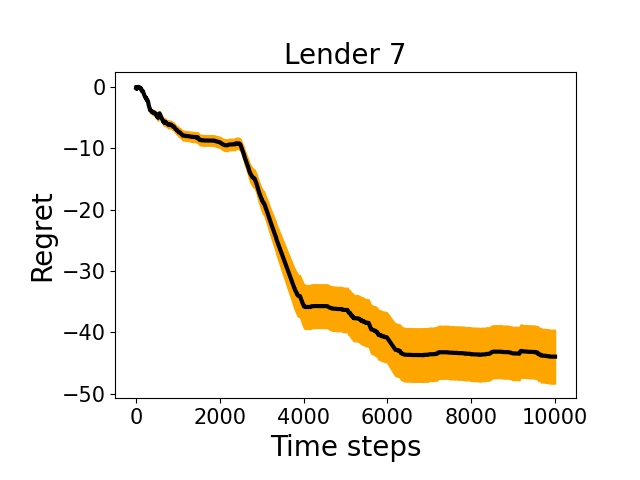}
\endminipage
\hfill
\caption{Cumulative expected lender regret over time simulated over 50 runs.}
\label{fig:basic_regrets}
\end{figure}
for lenders 53 and 37 shown in Figure~\ref{fig:basic_regrets}, we find that the regret either comes to near equilibrium or moves slowly towards saturation. However, for lenders where we find the regret becomes stationary like lender 37 after certain time steps, the learning rate towards better choices can be slower than other cases as the regret shown for lender 53. We find that lenders like lender 50 are not able to learn as best as the \textit{optimal solution} in that we see the regret is non-decreasing over the course of the simulation. However, what is interesting to see is that for lender 7, we see the regret is non-increasing, meaning that the lender utilities of lender 7 for all time steps are at least as good as the one returned by the \textit{optimal solution}. So in the presence of complicated functions characterizing the lender returns, optimizing the lender and borrower utilities as the matching objective as done for the \textbf{optimal solution} might not always yield the best lender returns.
\begin{figure}[!h]
\centering
\minipage{0.25\textwidth}
\includegraphics[width=4.5cm, height=3cm]{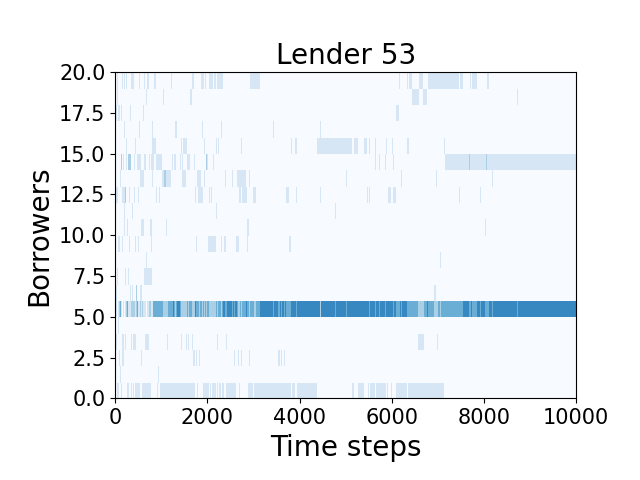}
\endminipage
\minipage{0.3\textwidth}
\includegraphics[width=4.5cm, height=3cm]{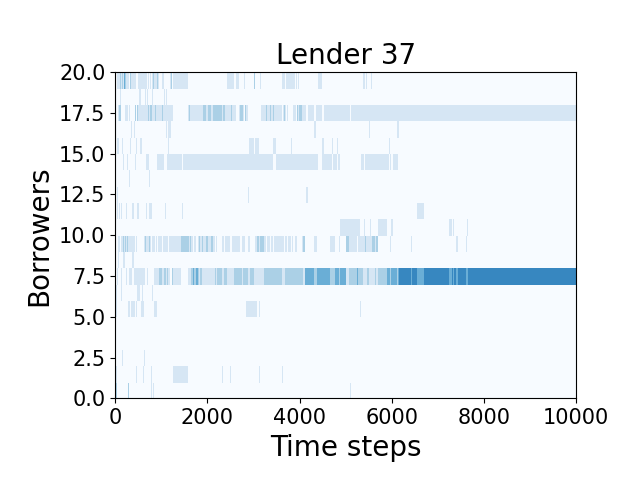}
\endminipage
\hfill
\\
\minipage{0.25\textwidth}
\includegraphics[width=4.5cm, height=3cm]{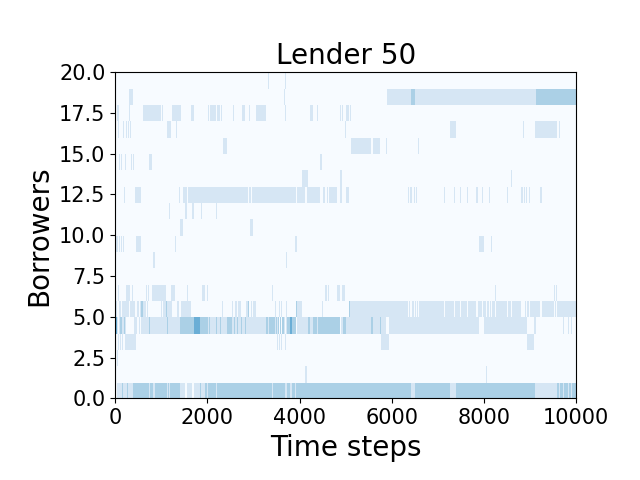}
\endminipage
\minipage{0.3\textwidth}
\includegraphics[width=4.5cm, height=3cm]{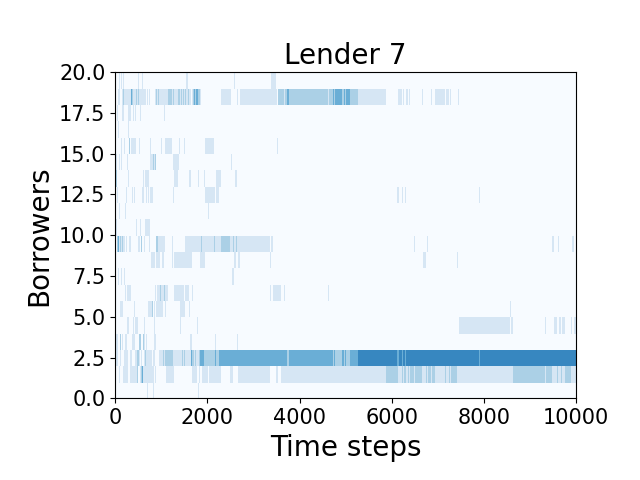}
\endminipage
\hfill
\caption{Heatmaps showing the lender matchings over the time steps from 50 simulations.}
\label{fig:basic_hmap}
\end{figure}

When we look at the heatmaps in Figure~\ref{fig:basic_hmap} where the grids denote the count of the number of times the lender was matched to borrower arm at each time step over 50 simulation runs. When considering the regret plots in Figure~\ref{fig:basic_regrets} with these heatmaps, we find that despite the stochasticity of the rewards, the lenders 53 and 37 start getting matched to the same borrower at the time when their regret starts to saturate. For lender 50, we find that the lender is not able to learn the best arm over the horizon which also explains its increasing regret over time. Note that the increasing regret dynamic can be explained by  two situations: one when the lender has learnt the wrong optimal arm and when the lender is not able to learn any one arm over time steps.

We next look at the comparison of the 3 settings: GS-UCB, GS-BLEMET, GS-BLEMET-FAIR described in Section~\ref{sec:seq_dec} to understand how the dynamics of cumulative regret compare among these methods. We want to outright mention that although the algorithm GS-BLEMET uses the UCB algorithm for deciding the termination thresholds, the cumulative regret of some lenders would be constant after they have been terminated from the matching although the matching for other lenders continue. 
\begin{figure}[!h]
\centering
\minipage{0.25\textwidth}
\includegraphics[width=4.5cm, height=3cm]{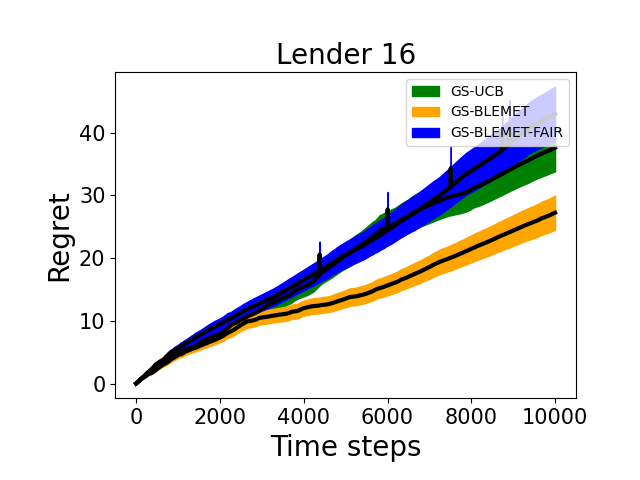}
\endminipage
\minipage{0.3\textwidth}
\includegraphics[width=4.5cm, height=3cm]{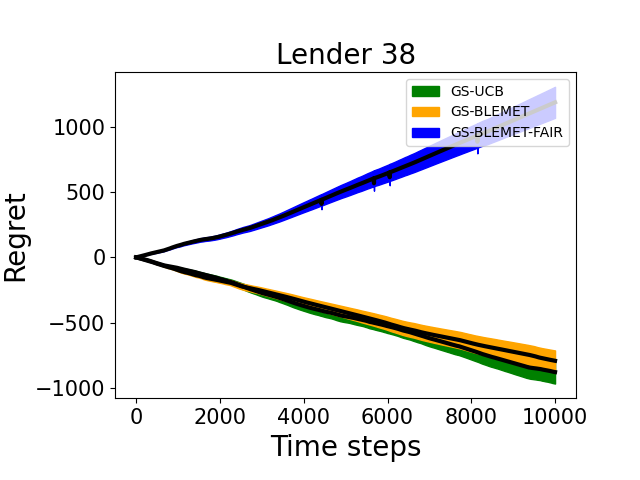}
\endminipage
\hfill
\\
\minipage{0.25\textwidth}
\includegraphics[width=4.5cm, height=3cm]{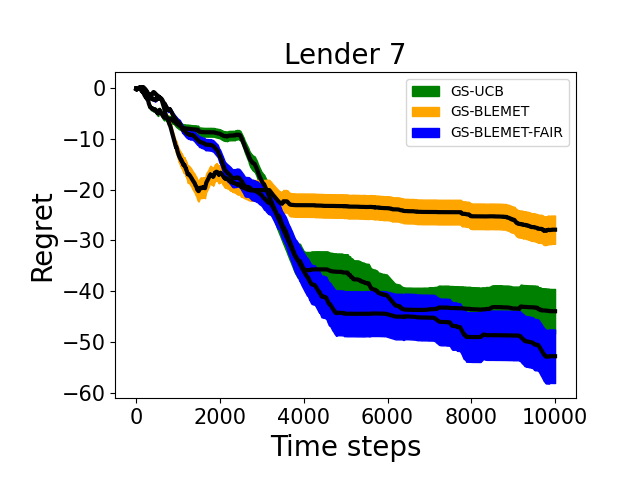}
\endminipage
\minipage{0.3\textwidth}
\includegraphics[width=4.5cm, height=3cm]{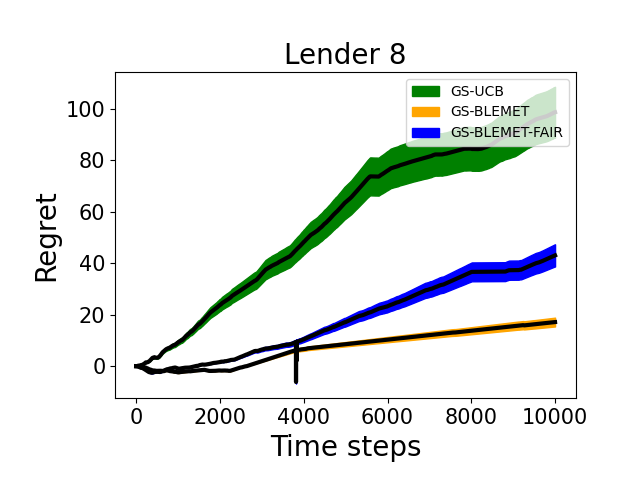}
\endminipage
\hfill
\caption{Cumulative expected lender regret over time simulated over 50 runs.}
\label{fig:regret_blemet}
\end{figure}
From Figure~\ref{fig:regret_blemet}, we find that the lender regrets can be categorize into three groups. First, represented in the regret plots for lenders 16 and 38, we find the GS-BLEMET regret is as good as the GS-UCB regret over all time steps and also better for lender 38. However,  in these cases we find that 
the fairness constraint using the maximum remaining amount for a borrower to penalize the objective in GS-BLEMET-FAIR hurts the regret of these lenders compared to the other algorithms. On the other hand, we find that for lenders 7 and 8, the cumulative regret is better for GS-BLEMET-FAIR than GS-UCB, both when the cumulative regrest either increase or decrease over time. These aggregated plots attest to the observation that there is not one rule that the dynamics of regret follows especially as the matching objective and constraints include a lot of factors into consideration. Recall that the main objective for GS-BLEMET-FAIR is to improve overall borrower regrets for the worst performing borrower, despite this we observe that the lender regrets although impacted for some lenders, are not worse according to this algorithm for every lender. 
\begin{figure}[!h]
\centering
\includegraphics[width=6cm, height=4cm]{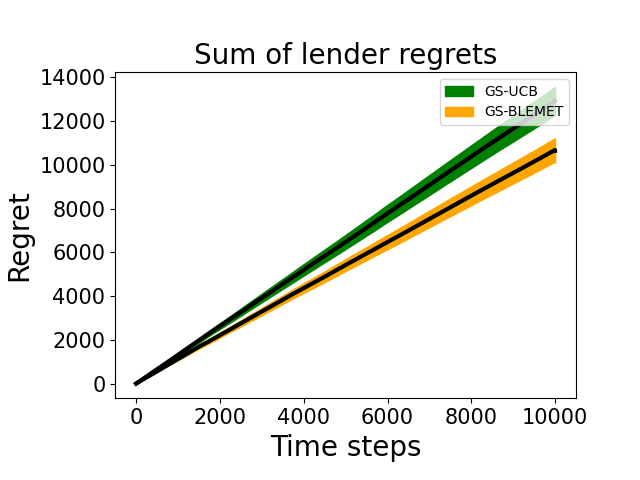}
\caption{Sum of the regrets over all lenders over time simulated over 50 runs.}
\label{fig:regret_sum}
\end{figure}
However, when we compare the sum of the lender regrets for both GS-UCB and the GS-BLEMET algorithms, we do find that the early termination criteria helps in lowering the cumulative regret over time as shown in Figure~\ref{fig:regret_sum}.

\begin{figure}[!t]
\centering
\includegraphics[width=6cm, height=4cm]{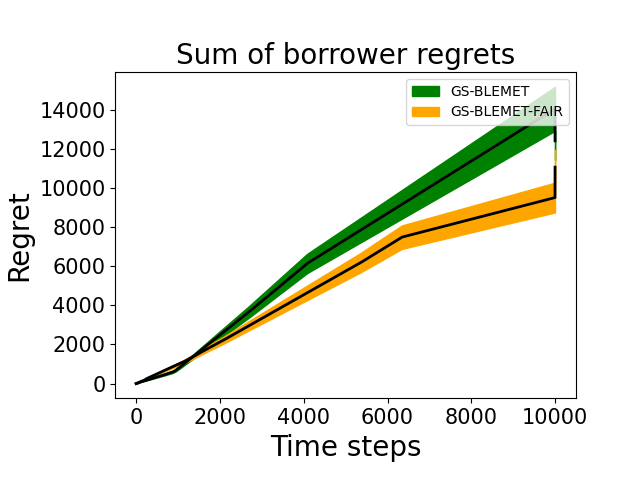}
\caption{Sum of the regrets over all borrowers over time simulated over 50 runs.}
\label{fig:regret_borrower}
\end{figure}

Finally, to analyze the effect of the fairness constraint on borrower side regret, we compute the borrower regret at $t$ as the following: $t* \sum_{l \in \mathcal{M}_{b, opt}} u_b(l) - \sum_{i=1}^t \sum_{l', \in \mathcal{M}_{b, alg}} \mathbb{E} \ \mathcal{R}(u_{b}(l'))$. This computes the regret from the sum of the optimally matched lenders in $\mathcal{M}_{b, opt}$ and the matched lenders from Algorithm~\ref{alg:blemet} given by $\mathcal{M}_{b, alg}$. When we see the plot in Figure~\ref{fig:regret_borrower}, we find that the algorithm GS-BLEMET with added fairness constraints in Equation~\ref{eq:opt_fair} has cumulative regret for the borrowers that scales lower than the optimization objective in GS-BLEMET. What is interesting to note is that although we do not optimize for the borrower utilities in the fairness constraints, but on the monetary budgets of the borrowers, this indirectly leads to improved borrower regrets over time considering the sum over all the borrowers, an observation we intend to explore in future research studies. However, what it suggests is that the sum of the borrower utilities are a surrogate to the monetary constraints in the framework of our matching objective in Equation~\ref{eq:gs_opt} and it also suggests that optimizing for the borrower equitable allocations can also result in the improvement in the overall regret considering the dynamics of time.

\section{Conclusions and Future work}
We consider a case of a centralized matching platform which requests proposals from borrowers about their preferences over the lenders or the agents through their personal rankings of the lenders. Then the platform decides the matching by allowing the lenders to interact with the platform over multiple time steps by either  accepting or rejecting the assigned borrower at a time step. From a lender perspective, this schema thus allows them to get matched without having the information of the actual returns while allowing them certain flexibility to exploit their options. The goal of this paper has been to lay out some ideas in which centralized peer lending platforms can be abstracted from a matching market perspective and how bandits could play a role in such mechanism design. These matching markets allow for more privacy as well as ensuring equitable outcomes and in our situation can be achieved by designing proper utility functions and rewards for the agents. Similarly, in future one could design decision making in which lenders can elicit information about their peer choices as well as from networks that have been known to aid funding situations \cite{horvat2015network} . In this paper, we did not consider externalities in the matching especially, when information among agents over time can change the dynamics of matching. Such effects of peer preferences\cite{bodine2011peer} or externalities\cite{pycia2019matching, mannor2011bandits} have been known to introduce substitutes in the markets. One of the tenets of crowdfunding platforms is the probability of the borrower listing getting fully funded and that additionally constitutes the uncertain elements at the start of the matching. In our settings, this constraint was not violated as we included it as part of our constraints explicitly. However, in real world, many projects will not reach their reserve prices for the requested loans and such cases need to be handled explicitly. One of the future studies extending our work is to look at the connections between the utilities derived from recommending borrowers to lenders explicitly and how the lenders' preference revisions using these recommendations can improve the learning over time. Such recommendations can also be modeled to ensure the overall system reaches an equilibrium with maximal social welfare.

\small

\bibliographystyle{ACM-Reference-Format}
\bibliography{sample-base}

\end{document}